\documentclass[10pt,journal,compsoc]{IEEEtran}

\ifCLASSOPTIONcompsoc
  \usepackage[nocompress]{cite}
\else
  \usepackage{cite}
\fi

\usepackage{amsmath,amsfonts}
\usepackage{amssymb}
\usepackage{algorithmic}
\usepackage{algorithm}
\usepackage{array}
\usepackage[caption=false,font=normalsize,labelfont=sf,textfont=sf]{subfig}
\usepackage{textcomp}
\usepackage{stfloats}
\usepackage{url}
\usepackage{verbatim}
\usepackage{graphicx}
\usepackage{cite}
\usepackage{booktabs}
\usepackage{multirow}
\usepackage{graphicx}
\usepackage{xcolor}
\begin{document}
%
\title{Cascaded and Generalizable Neural Radiance Fields for Fast View Synthesis}
%
%
%
%

\author{Phong Nguyen-Ha, Lam Huynh, Esa Rahtu, Jiri Matas, Janne Heikkil\"a 
\IEEEcompsocitemizethanks{\IEEEcompsocthanksitem{Phong Nguyen-Ha, Lam Huynh and Janne Heikkil\"a are with the Center of Machine Vision and Signal Analysis, University of Oulu, Finland (e-mail: phong.nguyen@oulu.fi; lam.huynh@oulu.fi; janne.heikkila@oulu.fi).}

\IEEEcompsocthanksitem{Esa Rahtu is with the Faculty of Information Technology and Communication Sciences, Tampere University, Finland (e-mail: esa.rahtu@tuni.fi).}

\IEEEcompsocthanksitem{Jiri Matas is with the The Center for Machine Perception
Department of Cybernetics, Czech Technical University, Prague (e-mail: matas@fel.cvut.cz).}

}

}

%
%

\markboth{Journal of \LaTeX\ Class Files,~Vol.~14, No.~8, August~2015}%
{Shell \MakeLowercase{\textit{et al.}}: Bare Demo of IEEEtran.cls for Computer Society Journals}
%



\IEEEtitleabstractindextext{%
\begin{abstract}

We present CG-NeRF, a cascade and generalizable neural radiance fields method for view synthesis. Recent generalizing view synthesis methods can render high-quality novel views using a set of nearby input views. However, the rendering speed is still slow due to the nature of uniformly-point sampling of neural radiance fields. Existing scene-specific methods can train and render novel views efficiently but can not generalize to unseen data. Our approach addresses the problems of fast and generalizing view synthesis by proposing two novel modules: a coarse radiance fields predictor and a convolutional-based neural renderer. This architecture infers consistent scene geometry based on the implicit neural fields and renders new views efficiently using a single GPU. We first train CG-NeRF on multiple 3D scenes of the DTU dataset, and the network can produce high-quality and accurate novel views on unseen real and synthetic data using only photometric losses. Moreover, our method can leverage a denser set of reference images of a single scene to produce accurate novel views without relying on additional explicit representations and still maintains the high-speed rendering of the pre-trained model. Experimental results show that CG-NeRF outperforms state-of-the-art generalizable neural rendering methods on various synthetic and real datasets.

\end{abstract}

}

\maketitle

\IEEEdisplaynontitleabstractindextext

%
\IEEEpeerreviewmaketitle

\section{Introduction}
Novel view synthesis (NVS) is a long-standing task in computer vision and computer graphics that has applications in free-viewpoint video, telepresence, and mixed reality~\cite{VR}. Novel view synthesis is a problem where visual content is captured from a set of sparse reference views and synthesized for an unseen target view. 
The problem is challenging since mapping between views depends on the 3D geometry of the scene, and the camera poses between the views. Moreover, NVS requires not only the propagation of information between the views but also the hallucination of details in the target view that is not visible in the reference image due to occlusions or limited field of view.

Early NVS methods produced target views by interpolating in ray \cite{intro_ray} or pixel space \cite{intro_pixel}. 
They were followed by works that leveraged geometric constraints such as epipolar consistency \cite{intro_warp} for depth-aware warping of the input views. These interpolation-based methods suffered from artifacts arising from occlusions and inaccurate geometry.
Later works tried to patch the artifacts by propagating depth values to similar pixels \cite{intro_similardepth} or by soft 3D reconstruction \cite{intro_soft3d}. 
However, these approaches cannot leverage depth information to refine the synthesized images or deal with the unavoidable issues of temporal inconsistency.
Recently, Neural Radiance Fields (NeRF) significantly impacted NVS research by implicitly representing the 3D structure of the scene and rendering novel photorealistic images. There are two main drawbacks of NeRF~\cite{Nerf}: i) the requirement to train from scratch for every new scene separately and ii) slow rendering speed. Moreover, the per-scene optimization of NeRF is lengthy and requires densely captured images for each scene.

Recent approaches~\cite{yu2020pixelnerf, trevithick2021grf, chen2021mvsnerf,chibane2021stereo,geonerf, pointnerf} address the former issue by training a generalized NeRF model to unseen scenes. The standard strategy is to condition the NeRF renderer with features extracted from source images from nearby views. Despite the generalization ability of these models to new scenes, the rendering speed is a bottleneck, and they cannot render novel views at an interactive rate. 
Chen et al.~\cite{chen2021mvsnerf} decodes multi-view input features by using time-consuming 3D convolution and  Multi-Layer Perceptron (MLP) networks into volume densities and radiance colors of the high-resolution target images.
Rendering such images requires querying millions of input 3D points to the model, so it is non-trivial to render the entire novel views in a single forward pass. There are recent scene-specific NeRF-based methods that can render photorealistic novel images in real-time and require less than an hour for training. Despite such impressive results, these methods often rely on either differentiable explicit voxel representation~\cite{sparse_voxels,Plenoxels,sun2021direct} or a multi-resolution hash table~\cite{mueller2022instant} to store the neural scene representation. Therefore, those methods require a completely new per-scene optimization step to render novel views of an unseen data.

This work addresses the above issues by proposing a novel and efficient view synthesis pipeline that renders the entire view in a single forward pass during training and testing. Inspired by the recently proposed works~\cite{nguyen2021rgbd}, we
adopt the coarse-to-fine RGB and depth rendering scheme to speed up the rendering process. 
Similar to MVSNeRF~\cite{chen2021mvsnerf}, we also infer a low-resolution 3D volume from a few unstructured multi-view input images using a shallow, yet efficient attention-based network. 
We found that synthesizing a low-resolution novel view using NeRF is fast and efficient due to the reduced amount of sampled 3D points. Moreover, the volume rendering of NeRF provides low-resolution radiance features and depth maps at the novel viewpoints. Instead of using the time-consuming coarse-to-fine rendering approach like~\cite{Nerf,yu2020pixelnerf}, we use the inferred depth maps to produce near-depth features of the target viewpoint and then fuse them with the radiance features as inputs to a convolutional-based neural renderer. We then train both networks to esitmate high-resolution target images from low-resolution radiance features.
Rendering the entire novel view also allows us to use perceptual loss~\cite{vggLoss} or adversarial training~\cite{eff3DGAN}, enhancing the generated images' overall quality. We also include a regularization loss to ensure the predicted final images are consistent with the coarse estimated novel views from the coarse radiance field predictor.

\begin{figure*}[t]
\centering
  \includegraphics[width=0.99\linewidth]{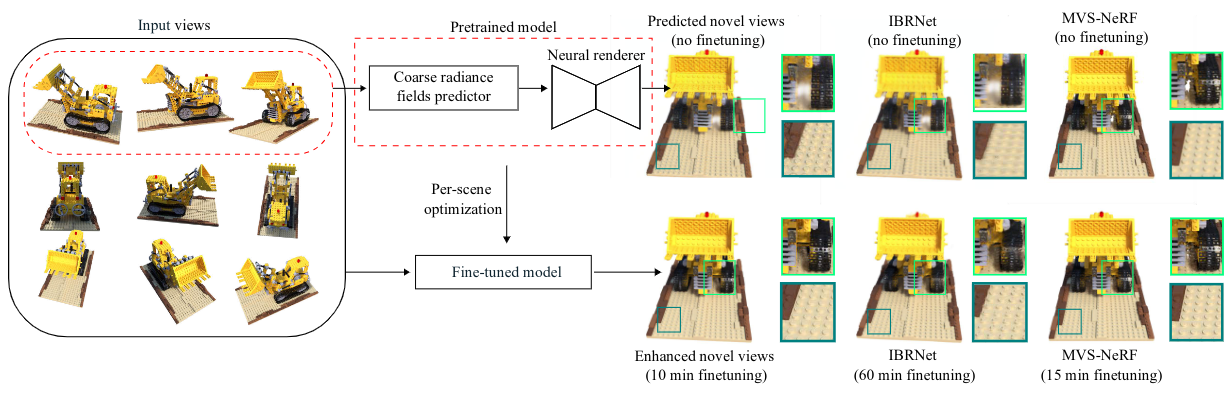}
  \caption{CG-NeRF is an efficient sparse view synthesis network that estimates coarse neural radiance fields of the target viewpoints at a low resolution and then renders the \textit{entire} novel images efficiently via a convolutional-based neural renderer. 
  Previous works~\cite{wang2021ibrnet,chen2021mvsnerf} do not possess such a renderer but rely on a deep fully connected network to estimate the high-resolution novel images pixel-by-pixel. Therefore, multiple forward passes are required to render pixels of novel views and this process often takes minutes to finish. In contrast, CG-NeRF renders coarseally the \textit{entire} novel view more efficiently in a single forward pass and also requires less fine-tuning time than previous works to achieve state-of-the-art results.}
\label{fig_overview}
\end{figure*}

Our trained CG-NeRF model renders plausible results on target poses close to the input viewpoints. However, the performance degrades when we extrapolate the targets further from the nearby source views.
Previous works~\cite{chen2021mvsnerf,sun2021direct,Plenoxels,sparse_voxels} learn a hybrid implicit-explicit representation of radiance fields using a denser set of input images that cover more views of a single scene.
Using a similar approach, finetuning the pre-trained CG-NeRF model in 10-15 minutes produces state-of-the-art results compared to those produced by scene-specific approaches~\cite{pointnerf,sun2021direct}. Both pre-trained and finetuned CG-NeRF models do not require explicit data structure but rely on a few selected reference views closest to the targets.
As can be seen in Fig.~\ref{fig_overview}, we also observe clear improvements in terms of visual quality between the novel views generated by the CG-NeRF models and the other view synthesis methods~\cite{yu2020pixelnerf,chen2021mvsnerf,wang2021ibrnet}. Note that our method does not rely on depth supervision~\cite{nguyen2021rgbd} to improve the quality of the synthesized images.

CG-NeRF shows strong generalizability to render realistic images at novel viewpoints via a lightweight view synthesis network. If shortly further optimized with additional images, CG-NeRF outperforms both recently proposed generalizable view synthesis methods~\cite{chen2021mvsnerf,pointnerf,wang2021ibrnet} and per-scene optimized models~\cite{Nerf,sun2021direct}. The main contributions of the work are:
\begin{itemize}
    \item An efficient sparse view synthesis network that employs a coarse radiance field predictor and a neural renderer to effectively predict novel images approximately two orders of magnitude faster than NeRF~\cite{Nerf} and its variants~\cite{chen2021mvsnerf,pointnerf,wang2021ibrnet}.
    \item The proposed scene-specific model requires only 10-15 minutes of fine-tuning the pre-trained model using more images. In addition, CG-NeRF does not require additional depth supervision.
    \item CG-NeRF achieves state-of-the-art results in novel view synthesis on both real and synthetic datasets such as DTU~\cite{DTU}, Synthetic-NeRF~\cite{Nerf}, Forward-Facing~\cite{llff} and Tank\&Temples~\cite{TankAndTemples}.
\end{itemize}
We will publicly publish the source code and neural network models upon paper publication.

\section{Related Works}
In the following, we discuss different generalizable view synthesis methods using a set of sparse input views. We would like to refer to \cite{report, xie2021neuralfield} for a more extensive review.

\noindent{\textbf{Novel view synthesis.}} Early works based on deep learning often use a Plane Sweep Volume (PSV) \cite{PSV}.
Each input image is projected onto successive virtual planes of the target camera to form a PSV. Kalantari et al. \cite{Kalantari} calculate the mean and standard deviation per plane of the PSV to estimate the disparity map and render the target view. Extreme View Synthesis (EVS) \cite{EVS} builds upon DeepMVS \cite{DeepMVS} to estimate a depth probability volume for each input view that is then warped and fused into the target view. A similar coarse-to-fine scheme has been proposed by Nguyen et al. ~\cite{nguyen2021rgbd}, but the method relies on depth supervision for view synthesis. Rather than estimating the depth maps of the source images, we train CG-NeRF to predict the depth map at the target view via volumetric rendering. The inferred depth is then used to produce high-resolution appearance features, which are later rendered as novel views.

\noindent \textbf{Multi-layered representation.} A significant number of works \cite{StereoMag,MPI2,MPI3,deepview} on view synthesis represent the 3D scene by Multiple Plane Images (MPIs). Each MPI includes multiple RGB-$\alpha$ planes, where each plane is related to a certain depth. The target view is generated by using alpha composition \cite{alpha} in the back-to-front order. Zhou et al. \cite{StereoMag} introduce a deep convolutional neural network to predict MPIs that reconstruct the target views for the stereo magnification task.
Local Light Field Fusion (LLFF) \cite{llff} introduces a practical high-fidelity view synthesis model that blends neighboring MPIs to the target view. The input to the MPI-based methods is also PSVs. However, those PSVs are constructed for a fixed range of depth values. The proposed CG-NeRF leverages coarse geometry and gathers near-surface features to enrich fine PSVs.

\noindent \textbf{Voxel grid.} Grid-based representations are similar to the MPI representation but are based on a dense uniform grid of voxels. This representation has been used as the basis for neural rendering techniques to model object appearance. Neural Volumes~\cite{NV} is an approach for learning dynamic volumetric representations of multi-view data. The main limitation of grid-based methods is the required cubic memory footprint. The sparser the scene, the more voxels are empty, which wastes model capacity and limits output resolution. 
A recent work by Sun et al.~\cite{sun2021direct} represents a 3D scene using low-resolution density and feature voxel grids for scene geometry and appearance. This method is fast to fit and produces high-quality novel views comparable with our fine-tuned CG-NeRF model on a single scene. Instead of optimizing such voxel-grid representation~\cite{NV,Plenoxels,sun2021direct}, we propose using a memory-efficient architecture to encode multi-view input features into a single volume and infer the coarse geometry and appearance features of the entire target view in a single forward pass.

\noindent \textbf{Pointclouds.} Recent works~\cite{synsin,NPBG,Pulsar,stable} on view synthesis have also employed the point-based representation to model 3D scene appearance. A drawback of the point-based representation is that there might be holes between points after projection to the screen space. Aliev et al. \cite{NPBG} train a neural network to learn feature vectors that describe 3D points in a scene. These learned features are then projected onto the target view and fed to a rendering network to produce the final novel image. A recent work by Xu et al.~\cite{pointnerf} proposes a point-based radiance field representation that efficiently renders novel views within 15 minutes of training for each new scene. However, this method requires ground-truth depths to train a multi-view depth estimator network. In contrast, we only leverage photo-metric losses between the generated and ground-truth novel views to train our model. Experimental results show that CG-NeRF can produce temporally consistent depths and novel views between multiple target viewpoints without relying on 3D supervision.

\noindent \textbf{Neural radiance fields.} The current state-of-the-art method Neural Radiance Fields (NeRF) by Mildenhall et al. \cite{Nerf} represents the plenoptic function by a multi-layer perceptron that can be queried using classical volume rendering to produce novel images. NeRF has to be evaluated at many sample points along each camera ray. This makes rendering a full image with NeRF extremely slow. Despite the high quality of the synthesized novel images, NeRF also requires per-scene training. Recent volumetric approaches \cite{yu2020pixelnerf, trevithick2021grf, chen2021mvsnerf,chibane2021stereo,geonerf, pointnerf} address the generalization issue of NeRF by incorporating a latent vector extracted from reference views. These methods show generalizability on selected testing scenes, but they share the slow rendering property of NeRF \cite{Nerf}. 
There are recent approaches that address the slow rendering of NeRF by sampling a
chunk of rays in a local patch and applying ConvNets for
post-processing such as enhancements or super-resolution. Despite having impressive results, those methods~\cite{AligNeRF,huang2023refsr} focus on rendering high resolution images via per-scene optimization.
In this work, we propose a generalizable view synthesis network which speeds up volume rendering with convolutional layers to estimate the target views efficiently on a single GPU. The  view synthesis results can also be enhanced by fine-tuning the obtained model on a single scene without using any additional components.

\section{Proposed Method}
This section describes in detail the architecture of CG-NeRF, which consists of two modules: a coarse radiance field predictor (Section~\ref{coarse_net}) that produces geometry and appearance of the scene at the lower resolution, and a convolutional-based neural renderer (Section~\ref{fine_net}) that combines both coarse and refined features to produce the final target image at the original size. In addition, we discuss the loss functions to train the generalizable CG-NeRF model and then finetune it on a single scene (Section~\ref{loss}).

\begin{figure*}[ht]
\centering  \includegraphics[width=0.9\linewidth]{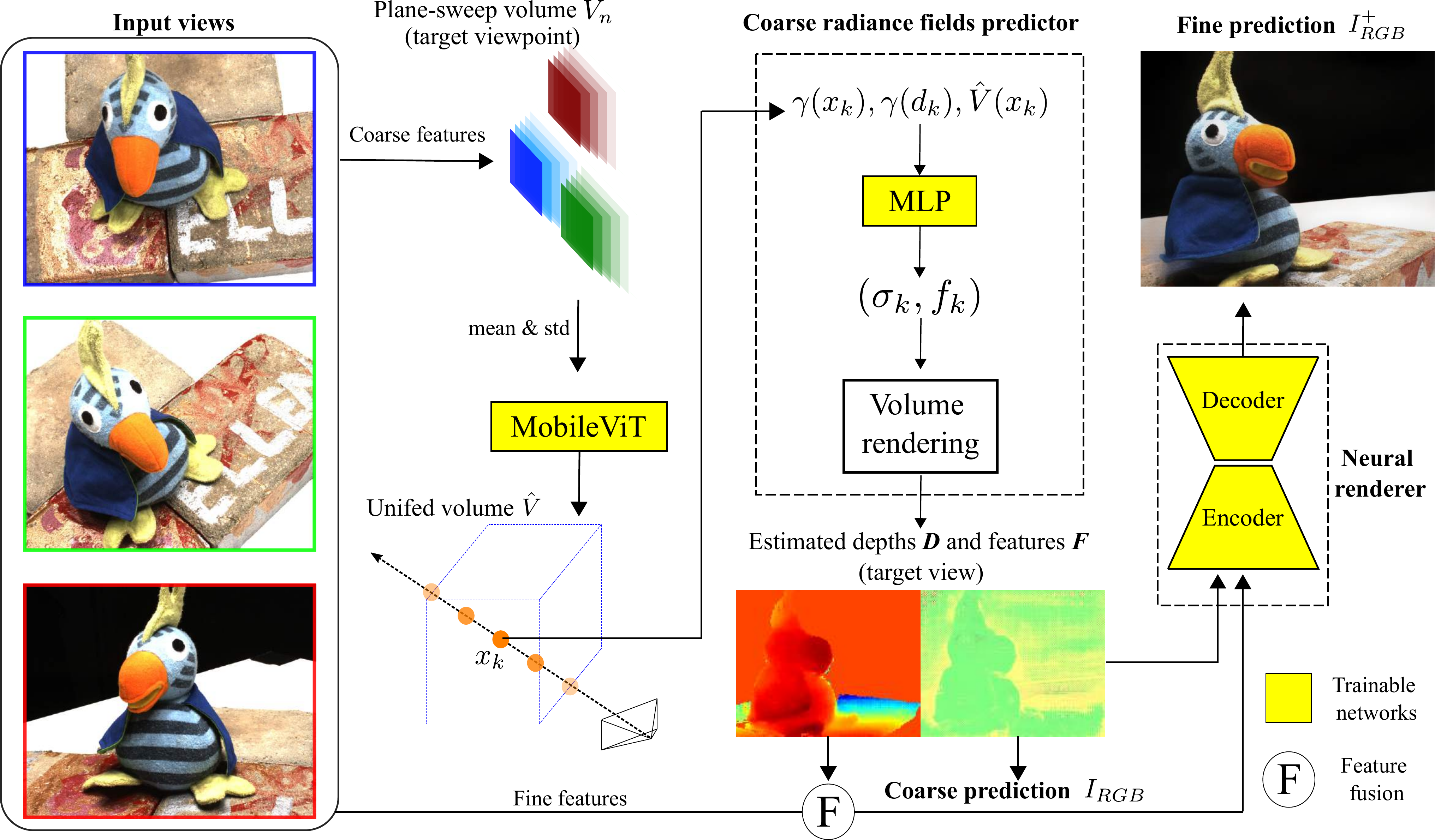}
  \caption{Our proposed CG-NeRF comprises several parts: (i) a memory-efficient MobileViT architecture~\cite{mehta2022mobilevit} that fused multiple low-resolution plane-sweep volumes of the target viewpoint into a single unified volume $\hat{V}$, (ii) a coarse radiance fields predictor that estimates target depth and features in low resolution, and (iii) an auto-encoder network to render novel views at the original resolution. Our method is lightweight and can infer fast novel views}.
\label{fig_coarse_net}
\end{figure*}


\subsection{Coarse radiance field predictor} \label{coarse_net}
Our approach to inferring coarse radiance fields is orthogonal to many recent works~\cite{chen2021mvsnerf,yu2020pixelnerf,wang2021ibrnet,chibane2021stereo} on generalized view synthesis. 
Despite impressive results, these methods cannot achieve fast view synthesis. Since each pixel is rendered independently, millions of query 3D points must pass through the deep networks. This scheme is expensive because the number of queries is much larger than the total number of pixels rendered.

The main difference between CG-NeRF and the above methods is that we infer radiance fields at a lower resolution to reduce queried inputs and speed up the rendering process. By doing so, we can also obtain the geometry and appearance features of the \textit{entire} target view in a single forward pass. Our method circumvents the slow rendering of NeRF~\cite{Nerf} by avoiding
splitting all queried 3D points into multiple chunks and rendering each chunk as a small image patch. Therefore, NeRF requires multiple forward passes to render patches of the novel views. This patch-based rendering strategy also prevents NeRF and its variants~\cite{chen2021mvsnerf,yu2020pixelnerf,wang2021ibrnet,chibane2021stereo} from training their model using GAN or perceptual losses~\cite{GAN,eff3DGAN} if stochastic pixels are generated during training. In contrast, we can train CG-NeRF on these losses between the ground-truth images and the estimated novel views at low and high resolution (see Section~\ref{loss}).

\noindent \textbf{Feature extraction}. We first describe our pipeline (see Fig.~\ref{fig_coarse_net}) for estimating coarse radiance fields of the target viewpoints given a set of $N$ unstructured input images $\{I_n\}_{n=1}^N$ and their poses.
Each input image $I_n$ is first fed to the Feature Pyramid Network \cite{FPN} to extract $F^c_n \in \mathbb{R}^{H/4 \times W/4 \times C}$ and $F^f_n \in \mathbb{R}^{H \times W \times C/4}$ which are coarse and fine 2D image features, respectively. Note that $F^f_n$ has the same height $H$ and width $W$ as the original input, so we can later use them for the coarse-to-fine synthesis. We do not know the scene geometry at the coarse level, so we uniformly sample several $K$ virtual depth planes. Therefore, we leverage the coarse features of each input view to build a cost volume at the target viewpoint. Those features $F^c_n$ are warped into multiple hypothesis depth planes via bilinear sampling~\cite{chen2021mvsnerf,nguyen2021rgbd}. The warped features are then concatenated to construct a per-view coarse volume $V_n \in \mathbb{R}^{H/4 \times W/4 \times K \times C}$.

\noindent \textbf{Multi-view attention learning}.
Each volume $V_n$ contains multiple-plane features of the target view so it requires a spatial reasoning architecture to aggregate those $N$ volumes before the neural rendering step. 
Previous works~\cite{wang2021ibrnet,nerformer} use the vanilla Transformers, which results in the slow inference of the novel views due to the heavy computation of multi-head attention~\cite{trasformer}. MVS-based methods~\cite{chen2021mvsnerf, nguyen2021rgbd} opt on the mean and variance-based volumes that a 3D UNet can later process to infer a unified scene encoding volume. However, a 3D Unet is limited due to the small receptive fields compared to attention-based architectures.

In this work, we combine the best of both approaches by using a single  MobileViT block~\cite{mehta2022mobilevit} which is a more memory-efficient variant of Transformers.
We also compute the mean and variance between $N$ volumes $V_n$ and concatenate them as a statistic volume which is then passed to a single MobileViT block.
This block learns the long-range dependencies via the multi-head attention~\cite{trasformer} between the non-overlapping patches of those $N$ volumes. We configure the input and output channels of the MobileViT block to produce a unified volume $\hat{V}$, which has the same spatial dimension as $V_n$. By learning to attend to extracted multi-view features, the inferred volume encodes both scene geometry and appearance, which can later be processed into volume densities and view-dependent features for view synthesis.

\begin{figure*}[ht]
\centering
  \includegraphics[width=0.95\linewidth]{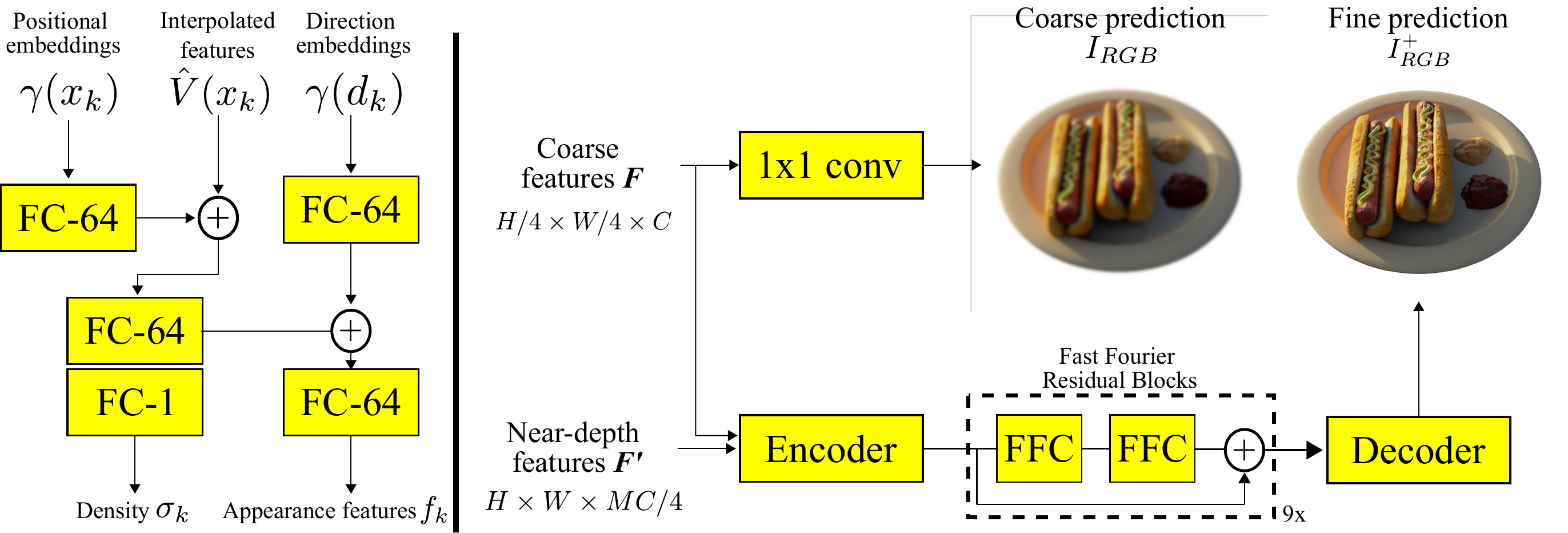}
  \caption{Our proposed MLP model (\textit{left}) uses several fully connected (FC) layers with 64 neurons each to estimate the coarse radiance fields of each sampled 3D point $x_k$. We also remove skip connections of the neural renderer (\textit{right}) and add nine residual blocks which utilize Fast Fourier Convolution (FFC)~\cite{FFC} layers for generating the fine prediction $I_{RGB}^+$ of the novel views. A 1x1 convolution layer can also be added to estimate a coarse prediction $I_{RGB}$ and regularize the training process. Finally, we use the dual discriminator of~\cite{eff3DGAN} to make sure that $I_{RGB}$ and $I_{RGB}^+$ are visually consistent with each other.}
\label{fig_arch}
\end{figure*}

\noindent \textbf{Coarse radiance fields}.
Using the unified coarse volume $\hat{V}$, our method learns an  MLP network $M$ (see Fig.~\ref{fig_arch} (left)) to regress volume density $\sigma_k \in \mathbb{R}^1$ and appearance feature $f_k\in \mathbb{R}^C$ of a 3D point $x_k\in \mathbb{R}^3$ and its viewing direction $d_k\in \mathbb{R}^3$ .
Specifically, each 3D point $x_k$ is the intersection between a ray shooting from the target camera and a virtual depth plane.
We obtain the feature $\hat{V}(x_k)$ of sampled point $x_k$ via trilinear interpolation~\cite{Occ_net}. The coarse radiance fields are then computed as follows:
\begin{equation}
\label{eqn_Unet}
(\sigma_k,f_k) = M(\gamma(x_k),\gamma(d_k),\hat{V}(x_k))
\end{equation}
where $\gamma$ is the positional encoding function~\cite{Nerf}.
In this work, we estimate the per-point features $f_k$, which can be used later
for the neural renderer (described in Section ~\ref{fine_net}). 

We design the model $M$ as a shallow MLP network so that the training/inference speed can be faster than its NeRF counterparts.
Instead of directly concatenating positional embedded $\gamma(x_k)$ and $\gamma(d_k)$ to image features, we use two different fully connected layers and project both embeddings into the latent space before combining them with the interpolated feature $\hat{V}(x_k)$. 
Adding those two extra layers does not increase the inference time but further improves the learning capacity of the model.
In addition, our proposed architecture runs more efficiently than its NeRF counterparts since it inherits the massive speedup of the fully-fused connected layers~\cite{mueller2022instant} by treating the entire network as a single GPU kernel.

We also adopt the volume rendering approach of~\cite{Nerf} to render novel depths and features via differentiable ray marching. Specifically, we can compute the per-ray feature $F_r$ by accumulating $f_k$ and $\sigma_k$ of $K$ sampled 3D points along a ray $r$ as follows:
\begin{equation}
\label{eqn_nerf_c}
F_r = \sum_K \tau_k(1-\exp(-\sigma_k\Delta_k))f_k
\end{equation}
\begin{equation}
\label{eqn_nerf_tau}
\tau_k = \exp(-\sum_{t=1}^{k-1}\sigma_t\Delta_t)
\end{equation}
where $\tau_i$ is the accumulated volume transmittance from ray origin to the point $x_k$ and $\Delta_k$ is the distance between adjacent sampling points. Gathering all ray-features $F_r$ of the target camera; we obtain the coarse feature maps $F \in \mathbb{R}^{H/4 \times W/4 \times C}$ of the novel view.
We can also obtain the temporally consistent depth maps $D \in \mathbb{R}^{H/4 \times W/4 }$ as a side-product of the volume rendering by calculating the weighted sum between estimated densities and depth values of $x_k$. Without learning to predict the novel depth maps at the original resolution, we up-sample the coarse depth maps $D$ and use them to combine high-resolution input features $F^f_n$ via a feature fusion step.

\subsection{Neural renderer} \label{fine_net}
\noindent \textbf{Feature fusion}. Once we get a coarse scene geometry and appearance estimation, we use an auto-encoder network to render the final target images. It is challenging to generate such high-resolution novel views using coarse features. Therefore, we leverage the depth plane resampling technique of~\cite{nguyen2021rgbd} to obtain near-depth features $F^\prime$, which have the exact spatial resolution as the target views.
For each ray shot from the target camera, we sample $J$ points around the predicted depth value of the given ray. We back-project those points into each input camera and obtain their corresponding features by bi-linearly interpolating the extracted high-resolution input features $F^f_n$. 
Each pixel of $F^\prime \in \mathbb{R}^{H \times W \times JC/4}$ is the weighted sum of the multi-view warped features, and weights are defined as the inverse depths of the $x_k$ at each input viewpoint. As the coarse radiance fields predictor improves at predicting depth maps, so does the feature fusion method.
Please refer to~\cite{nguyen2021rgbd} for more details.

\noindent \textbf{UNet renderer}. To render the novel views at the high resolution, we up-sample the coarse feature $F$ to match the spatial resolution of the original targets and then concatenate with $F^\prime$ before feeding to a fully convolutional UNet neural renderer which contains three up and down-sampling convolutional layers. Instead of using skip connections between the encoder and decoder, we employ several residual blocks with Fast Fourier Convolutions~\cite{FFC,Lama,HVS}. These blocks have the image-wide receptive field to effectively combine multi-scale features $F$ and $F^\prime$ to render the high-resolution novel images $I_{RGB}^+$. Without the skip connection between the encoder and decoder, the Unet model is also smaller and more efficient due to the reduced number of parameters.

\subsection{Loss functions} \label{loss}
We train both the coarse radiance field predictor and the auto-encoder network 
using a fine reconstruction loss $\mathcal{L}_{fine}$ which is a combination of the L1 loss and perceptual~\cite{vggLoss} loss between $I_{RGB}^+$ and the ground-truth images. 
A $1\times1$ convolutional layer can also be added to transform coarse features $F$ to low-resolution RGB color images $I_{RGB}$ (see Fig.~\ref{fig_arch} (right)). Outputing an intermediate coarse novel view allows us to train CG-NeRF using a reconstruction L1 loss term $\mathcal{L}_{coarse}$ between $I_{RGB}$ and the down-scale ground-truths. This loss term also helps to regularize the coarse radiance field learning without relying on the depth supervision~\cite{nguyen2021rgbd}. 
To make sure that both $I_{RGB}$ and $I_{RGB}^+$ are visually consistent with each other, we follow the dual discriminator set up of~\cite{eff3DGAN}, add a hinge GAN~\cite{GAN} loss $\mathcal{L}_{GAN}$. Instead of discriminating over three-channel images, we perform the adversarial training using six-channel real and fake images. The fake image $I_{RGB}$ is up-sampled before concatenating with $I_{RGB}^+$. The ground-truth image is also down-sampled before concatenating with the original image.
The total photometric loss to train CG-NeRF is computed as $\mathcal{L}_{total} = \mathcal{L}_{fine} + \mathcal{L}_{coarse} + \lambda \mathcal{L}_{GAN}$.

\begin{figure*}[t]
\centering
  \includegraphics[width=0.90\linewidth]{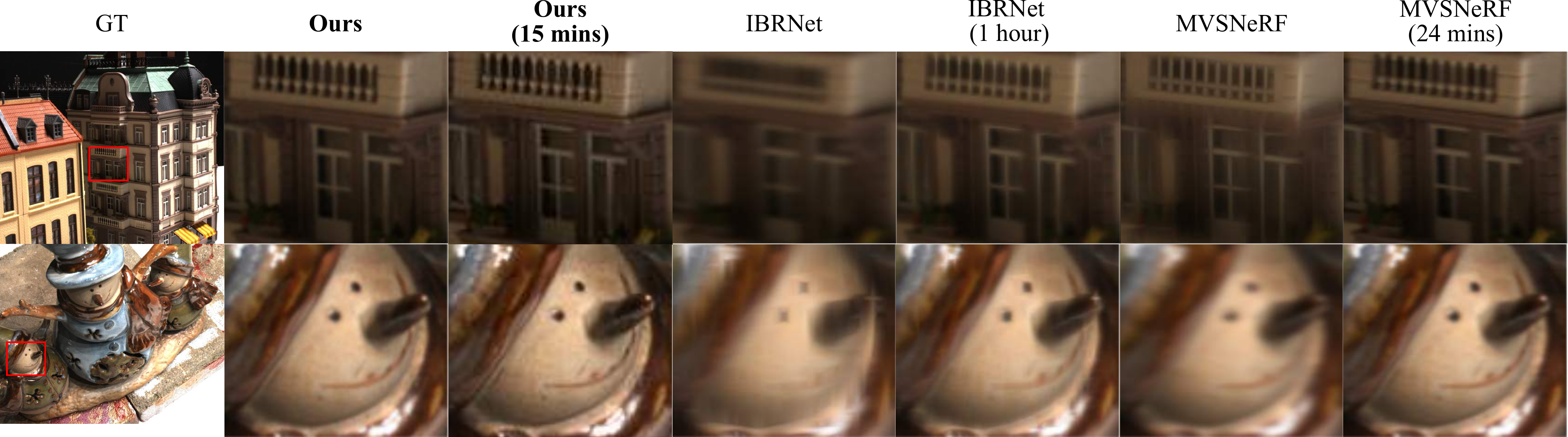}
  \caption{Qualitative comparisons of view synthesis on the two testing sets of the DTU dataset~\cite{DTU}. Our CG-NeRF can recover texture details and geometrical structures more accurately than other methods. Within 15 minutes of finetuning, CG-NeRF outperforms state-of-the-art methods such as MVSNeRF~\cite{chen2021mvsnerf} and IBRNet~\cite{wang2021ibrnet} which are required to more time to optimize using the same amount of data.}
\label{fig_DTU}
\end{figure*}

\begin{table*}[t]
\caption{Quantitative comparison on large-scale dataset of synthetic and real images.\\Methods with $\dagger$ and $*$ symbols are optimized per scene for 15 and 60 minutes respectively. 
}
\centering
\resizebox{0.99\textwidth}{!}{%
\begin{tabular}{@{}lcccccccccccc@{}}
\toprule
\multirow{2}{*}{Methods} &
  \multicolumn{3}{c}{DTU~\cite{DTU}} &
  \multicolumn{3}{c}{Synthetic-NeRF~\cite{Nerf}} &
  \multicolumn{3}{c}{Forward-Facing~\cite{llff}} &
  \multicolumn{3}{c}{Tanks \& Temples~\cite{TankAndTemples}} \\ \cmidrule(l){2-13} 
 &
  PSNR$\uparrow$ &
  SSIM$\uparrow$ &
  LPIPS$\downarrow$ &
  PSNR$\uparrow$ &
  SSIM$\uparrow$ &
  LPIPS$\downarrow$ &
  PSNR$\uparrow$ &
  SSIM$\uparrow$ &
  LPIPS$\downarrow$ &
  PSNR$\uparrow$ &
  SSIM$\uparrow$ &
  LPIPS$\downarrow$ \\ \midrule
pixelNeRF~\cite{yu2020pixelnerf} &
  19.31 &
  0.78 &
  0.38 &
  7.39 &
  0.65 &
  0.41 &
  11.24 &
  0.48 &
  0.67 &
  - &
  - &
  - \\
MVSNeRF~\cite{chen2021mvsnerf} &
  26.64 &
  0.93 &
  0.16 &
  23.62 &
  0.89 &
  0.17 &
  21.93 &
  0.79 &
  0.25 &
  23.46 &
  0.84 &
  0.26 \\
IBRNet~\cite{wang2021ibrnet} &
  26.04 &
  0.92 &
  0.19 &
  22.44 &
  0.87 &
  0.19 &
  21.79 &
  0.78 &
  0.28 &
  21.70 &
  0.81 &
  0.29 \\
PointNeRF~\cite{pointnerf} &
  23.89 &
  0.87 &
  0.2 &
  - &
  - &
  - &
  - &
  - &
  - &
  - &
  - &
  - \\
\textbf{CG-NeRF} &
  \textbf{28.21} &
  \textbf{0.93} &
  \textbf{0.17} &
  \textbf{25.01} &
  \textbf{0.90} &
  \textbf{0.19} &
  \textbf{23.93} &
  \textbf{0.82} &
  \textbf{0.21} &
  \textbf{28.12} &
  \textbf{0.88} &
  \textbf{0.16} \\ \midrule
$\text{MVSNeRF}^{\dagger}$~\cite{chen2021mvsnerf} &
  28.50 &
  0.94 &
  0.17 &
  27.05 &
  0.93 &
  0.16 &
  25.45 &
  0.87 &
  0.19 &
  28.39 &
  0.89 &
  0.15 \\
$\text{IBRNet}^{*}$~\cite{wang2021ibrnet} &
  31.35 &
  0.95 &
  0.13 &
  25.62 &
  0.94 &
  0.11 &
  24.88 &
  0.86 &
  0.18 &
  26.12 &
  0.85 &
  0.21 \\
$\text{PointNeRF}^{\dagger}$~\cite{pointnerf} &
  30.12 &
  0.95 &
  0.11 &
  30.71 &
  0.96 &
  0.08 &
  26.35 &
  0.88 &
  0.15 &
  29.61 &
  0.95 &
  0.08 \\
$\text{DVGO}^{\dagger}$~\cite{sun2021direct} &
  31.49 &
  0.96 &
  0.10 &
  31.95 &
  0.96 &
  0.05 &
  - &
  - &
  - &
  28.41 &
  0.91 &
  0.15 \\
Insant-NGP$^{\dagger}$~\cite{mueller2022instant} &
  32.49 &
  0.96 &
  0.08 &
  33.18 &
  0.96 &
  0.05 &
  26.14 &
  0.92 &
  0.13 &
  29.25 &
  0.94 &
  0.07 \\
\textbf{$\text{CG-NeRF}^{\dagger}$} &
  \textbf{33.52} &
  \textbf{0.98} &
  \textbf{0.08} &
  \textbf{34.12} &
  \textbf{0.98} &
  \textbf{0.04} &
  \textbf{27.32} &
  \textbf{0.93} &
  \textbf{0.13} &
  \textbf{30.15} &
  \textbf{0.96} &
  \textbf{0.06} \\ \bottomrule
\end{tabular}
}
\label{table_quant_NVS}
\end{table*}

\begin{figure*}[t]
\centering
  \includegraphics[width=0.99\linewidth]{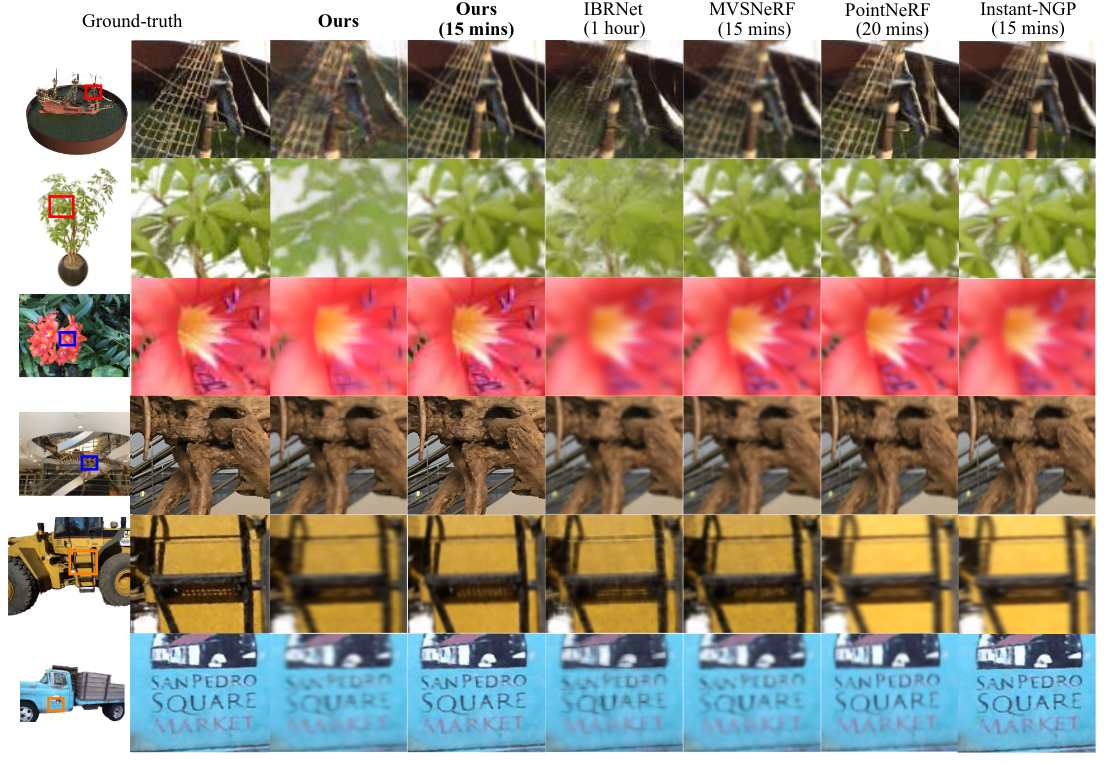}
  \caption{Qualitative comparisons between CG-NeRF and state-of-the-art methods on unseen data. 
  Each row show estimated novel views of three different datasets: Synthetic-NeRF~\cite{Nerf}, Forward-Facing~\cite{llff}, and Tanks \& Temples~\cite{TankAndTemples}. Without any finetuning, our pre-trained model can produce plausible results on unseen data and the results are vastly improved after optimizing on a single scene. In general, our method produces more accurate textures and thin structure objects than those produced by other baselines.}
\label{fig_Nerf}
\end{figure*}

\begin{table}[t]
\centering
\caption{Quantitative comparisons of rendering on 800x800 novel images between CG-NeRF and other view synthesis methods. Methods with $\dagger$ symbols are optimized for each testing scene.}
\begin{tabular}{@{}cccc@{}}
\toprule
Methods & Time$\downarrow$ & PSNR $\uparrow$ & \begin{tabular}[c]{@{}c@{}}Rendering time  $\downarrow$\\ (seconds)\end{tabular} \\ \midrule
pixelNeRF~\cite{yu2020pixelnerf} & 14 hours & 7.39 & 35 \\
IBRNet~\cite{wang2021ibrnet} & 1 days & 22.44 & 29 \\
MVSNeRF~\cite{chen2021mvsnerf} & 10 hours & 23.62 & 15 \\ \midrule
$\text{DVGO}^\dagger$~\cite{sun2021direct} & 15 minutes & 31.95 & 0.64 \\
$\text{PointNeRF}^\dagger$~\cite{pointnerf} & 15 minutes & 30.71 & 10 \\
$\text{Instant-NGP}^\dagger$~\cite{mueller2022instant} & 15 minutes & 33.18 & 0.65 \\
\textbf{\begin{tabular}[c]{@{}c@{}}$\text{CG-NeRF}^\dagger$\end{tabular}} & 15 minutes & \textbf{34.12} & \textbf{0.2} \\
\midrule
$\text{Instant-NGP}^\dagger$~\cite{mueller2022instant} & 5 minutes & 30.82 & 0.65 \\
\textbf{\begin{tabular}[c]{@{}c@{}}$\text{CG-NeRF}^\dagger$\end{tabular}} & 5 minutes & \textbf{31.25} & \textbf{0.2} \\
 \bottomrule
\end{tabular}
\label{tab_speed}
\end{table}

\begin{table}[t]
\centering
\caption{Quantitative comparisons of rendering on 800x800 and 4K novel images between CG-NeRF and other view synthesis methods using the Forward-Facing~\cite{llff} dataset. Methods with $\ddagger$ are optimized with the proposed neural renderer.}
\resizebox{0.50\textwidth}{!}{%
\begin{tabular}{@{}ccccc@{}}
\toprule
Resolution & Methods & Time$\downarrow$ & PSNR$\uparrow$ & \begin{tabular}[c]{@{}c@{}}Rendering time $\downarrow$ \\ (seconds)\end{tabular} \\ \midrule
\multirow{5}{*}{800x800} & pointNeRF~\cite{pointnerf} & 20 mins & 26.35 & 10 \\
 & $\text{pointNeRF}^\ddagger$~\cite{pointnerf} & 20 mins & 27.26 & 2.2 \\
 & Instant-NGP~\cite{mueller2022instant} & 15 mins & 26.14 & 0.65 \\
 & $\text{Instant-NGP}^\ddagger$~\cite{mueller2022instant} & 15 mins & 27.25 & 0.35 \\
 & $\textbf{CG-NeRF}^\ddagger$ & 15 mins & \textbf{27.32} & \textbf{0.2} \\ \midrule
\multirow{3}{*}{4K} & 
NeRF~\cite{Nerf} & 20 hours & 25.13 & 68 \\ &
$\text{NeRF}^\ddagger$~\cite{Nerf} & 20 hours & 25.86 & 20 \\
& Instant-NGP~\cite{mueller2022instant} & 1 hours & 26.45 & 19 \\
 & $\text{Instant-NGP}^\ddagger$~\cite{mueller2022instant} & 1 hours & 26.78 & 6 \\ 
 & $\textbf{CG-NeRF}^\ddagger$ & 1 hours & \textbf{27.12} & \textbf{2.5} \\ \bottomrule
\end{tabular}
}
\label{tab_improving}
\end{table}

\section{Experiments}
In this section, we evaluate CG-NeRF and compare the generated novel views to those produced by the  state-of-the-art methods.
\subsection{Implementations}

\noindent \textbf{Model details}.
The models were trained with the Adam optimizer using a 0.004 learning rate for the discriminator, 0.001 for both the coarse radiance field predictor and the neural renderer with the momentum parameters (0, 0.9). $\lambda= 0.5, C = 64, K = 64, J = 4, N = 3, W = 640, H = 512$ and the MobileViT block had 4 attention heads for fast inference. 
We used the Pytorch extension of the tiny-cuda-nn~\cite{tiny-cuda-nn} library for the fully-fused connected layers. The other modules of CG-NeRF are implemented using native PyTorch. We first trained the model from scratch on multiple scenes in 16 hours using four V100 GPUs with a batch size of 16. We then perform a 10-15 min fine-tuning on a single V100 GPU and achieve state-of-the-art results for per-scene optimization. We have tried to fine-tune the model on a consumer-grade RTX 2080TI GPU for 15 minutes and observe no significant differences in terms of view synthesis quality.

\noindent\textbf{View selection.} We follow the view selection method of~\cite{FVS} to choose the top 10 closest source images to each target image. We first run the standard structure-from-motion method~\cite{schoenberger2016sfm} on the each training scene to estimate the coarse depth map of each image in the dataset. We then project pixels of the novel views with valid depth values into each input image using the known camera intrinsics and extrinsics. For each novel view, we select the top 10 closest input images that have the most valid projections. Since our method is bounded by GPU memory, we then randomly sample $N$ out of 10 closest views. At each training step, the number of input views $N$ to the is further uniformly sampled between 3-5.

However, it is non-trivial to perform view extrapolation when the set of sparse $N$ near-by views can only cover a small area of the scene. We address this issue by increasing $N$ during the fine-tuning stage. We apply the same view selection strategy on other generalizing view synthesis baselines~\cite{wang2021ibrnet, chen2021mvsnerf} and we report the comparison results with them in the following sections. We also discuss more on how increase $N$ affects the training performance in the Table~\ref{table_inputs}.

\subsection{Experiments}

\noindent \textbf{Datasets}.
We train CG-NeRF on the DTU \cite{DTU} dataset to learn a generalizable network. DTU is an MVS dataset consisting of more than 100 scenes scanned in 7 different lighting conditions at 49 positions. From 49 camera poses, we selected 10 as targets for view synthesis and used the rest for source image selection.
We evaluate the performance of our pretrained model using the testing set of the DTU dataset. To further compare CG-NeRF with state-of-the-art methods, we test it on the Synthetic-NeRF~\cite{Nerf}, Forward-Facing~\cite{llff}, and Tanks \& Temples~\cite{TankAndTemples} datasets, which have different scenes and view distributions from our training set. Each scene includes 12 to 62 images and $1/8$ of these images are held out for testing. 

\noindent \textbf{Baselines}.
In this evaluation, we compare CG-NeRF to both generalizable view synthesis and pure per-scene optimized methods. The former approaches ~\cite{yu2020pixelnerf,chen2021mvsnerf,wang2021ibrnet} predict new views with and without per-scene optimization. We use their provided code and train them on DTU dataset~\cite{DTU} and then finetune for each testing scene for fair comparisons. 
Furthermore, we compare our method with the recent scene-specific synthesis methods such as Instant-NGP~\cite{mueller2022instant}, PointNeRF~\cite{pointnerf} and DVGO~\cite{sun2021direct}. We use their public code to train scene-specific models and compare the generated novel views with those produced by our method both qualitatively and quantitatively. We use a system of four V100 GPU to train and test all baselines and compare with our method for a fair comparison.

\noindent \textbf{Metrics}.
We report the PSNR, SSIM, and perceptual similarity (LPIPS) \cite{LPIPS} for CG-NeRF and other state-of-the-art methods. We summarize the quantitative and qualitative results in Table~\ref{table_quant_NVS}, Fig.~\ref{fig_DTU}, and Fig.~\ref{fig_Nerf} using samples from four different datasets~\cite{DTU,Nerf,llff,TankAndTemples}. Please see the supplementary video for more qualitative results.

\noindent \textbf{Testing on the seen dataset}.
We first evaluate CG-NeRF on the testing set of the DTU dataset.
Since our methods are trained on the training set of the same dataset, we observe both pretrained CG-NeRF and per-scene optimized {$\text{CG-NeRF}^\dagger$} can reconstruct accurate novel views on the unseen testing scenes as can be seen in the second and third column of Fig.~\ref{fig_DTU}. 
Moreover, they outperform other state-of-the-art methods both quantitatively and qualitatively. 
The direct inference network of IBRNet~\cite{wang2021ibrnet} and MVSNeRF~\cite{chen2021mvsnerf} are not able to produce faithful textures of the windows and the reflection at the tip of the nose of the toy character. Both baselines tend to predict blurry results and fail to retrieve fine details as can be seen in the zoomed insets of Fig.~\ref{fig_DTU}.
Since these methods predict each pixel of the high-resolution novel views from a low-resolution feature volume, their MLP network have to solve two difficult tasks which are image synthesis and super-resolution. Instead, we tackle these two tasks by two different networks: one to estimate coarse radiance features and another to refine them using the adversarial training and a convolution-based neural renderer. As can be seen in Fig.~\ref{fig_DTU}, 
the rendering results of $\text{IBRNet}^{*}$ and $\text{MVSNeRF}^{\dagger}$ are sharper if both models are optimized for each testing scene. Note that, they both required approximately 24 minutes up to 1 hour to achieve sharp results but still achieve less accurate novel views than ours which are fine-tuned in 15 minutes for all testing scenes.

\noindent \textbf{Testing on the unseen dataset}.
To further test the generalizability of our approach on unseen data, we conduct experiments on three synthetic and real datasets. As can be seen in the second column of Fig.~\ref{fig_Nerf}, CG-NeRF can produce plausible results on all testing novel views which are very different from the training DTU images. Despite not seeing those testing images, the CG-NeRF model shows competitive results with a variant of $\text{MVSNeRF}^{\dagger}$ which is trained for the same amount of time as ours. Given almost one hour of finetuning, $\text{IBRNet}^{*}$ improves its results significantly but is still not able to estimate as accurate novel views as ours. Optimizing for a quarter of an hour, $\text{CG-NeRF}^\dagger$ produces cleaner and more photo-realistic novel views than those produced by $\text{MVSNeRF}^{\dagger}$ and $\text{IBRNet}^{*}$ on these unseen synthetic and real datasets. 
Despite being trained on the testing scenes, both baseline methods are not able to render such fine details because they only use a L2 color loss between stochastic rendered and ground-truth pixels.
On the ship scene of the Synthetic-NeRF dataset, our optimized $\text{CG-NeRF}^\dagger$ model can render the thin structure of the sky-sail which is not visible on the generated novel views of other methods. 
High-frequency details can be seen in the generated images of our method on the real flower and trex scenes of the Forward-Facing dataset.
Moreover, we can also observe that $\text{CG-NeRF}^\dagger$ estimates clearer text on the door of the truck scene in the second example of the Tank\&Temples dataset.
Although our pretrained CG-NeRF model is not able to retrieve such fine details, the rendering results can be vastly improved thanks to the hinge GAN loss $\mathcal{L}_{GAN}$ that we applied during training. Note that, it is not straightforward for other baselines to  perform adversarial training due to the limited resolution of the generated images. This is not a problem with our methods because we can efficiently render the entire novel views without worrying about the out-of-memory issue which is unavoidable for other baselines.

We also compare our methods with state-of-the-art scene-specific view synthesis PointNeRF~\cite{pointnerf} models. Since this method is not designed for generalizable view synthesis, it performs worse than CG-NeRF. However, we observe a significant gain in performance of the fine-tuned $\text{PointNeRF}^{\dagger}$ model across testing data. As can be seen in the last column of Fig.~\ref{fig_Nerf}, $\text{PointNeRF}^{\dagger}$ can render fine details but the generated novel views are still not as accurate as ours in all testing scenes. In the challenging ficus scene of the Synthetic-NeRF dataset, our method can render the leaves sharper than $\text{PointNeRF}^{\dagger}$, which renders high-resolution novel views from a point-cloud where the feature of each point is interpolated from low-resolution image features. Despite using a memory-expensive point-cloud representation, $\text{PointNeRF}^{\dagger}$ is still not able to render high-quality details of the novel views, especially when we zoom closely into the content of the generated images.

\noindent \textbf{Rendering speed}.
In this section, we compare the rendering speed between the full CG-NeRF model and other view synthesis methods. In general, our method not only produces better novel views but also renders them faster than previous works.
Both pixelNeRF~\cite{yu2020pixelnerf} and IBRNet~\cite{wang2021ibrnet} takes more than half a minute to render a single novel image because the method uses the time-consuming MLP-based architecture for multi-view aggregation and it also inherits the slow rendering of NeRF. Moreover, it also takes several hours for training and still perform worse than our approach.

The point-based approach PointNeRF~\cite{pointnerf} improves the speed by rendering the novel views directly from their hybrid implicit-explicit volume representations. However, the method 
is still slow and not able to render novel views at the interactive rate. 
As can be seen in the Table~\ref{tab_speed}, our 5-15 minutes-finetuning CG-NeRF model not only outperforms the state-of-the-art scene-specific fast view synthesis methods~\cite{sun2021direct,mueller2022instant} but also renders novel at least 3 times faster than them. We found that rendering the entire novel views using the proposed fully-fused MLP and convolution-based neural renderer is faster than sequentially rendering individual pixels using a deep MLP model of NeRF and its variants.

\begin{figure*}[t]
\centering
  \includegraphics[width=0.9\linewidth]{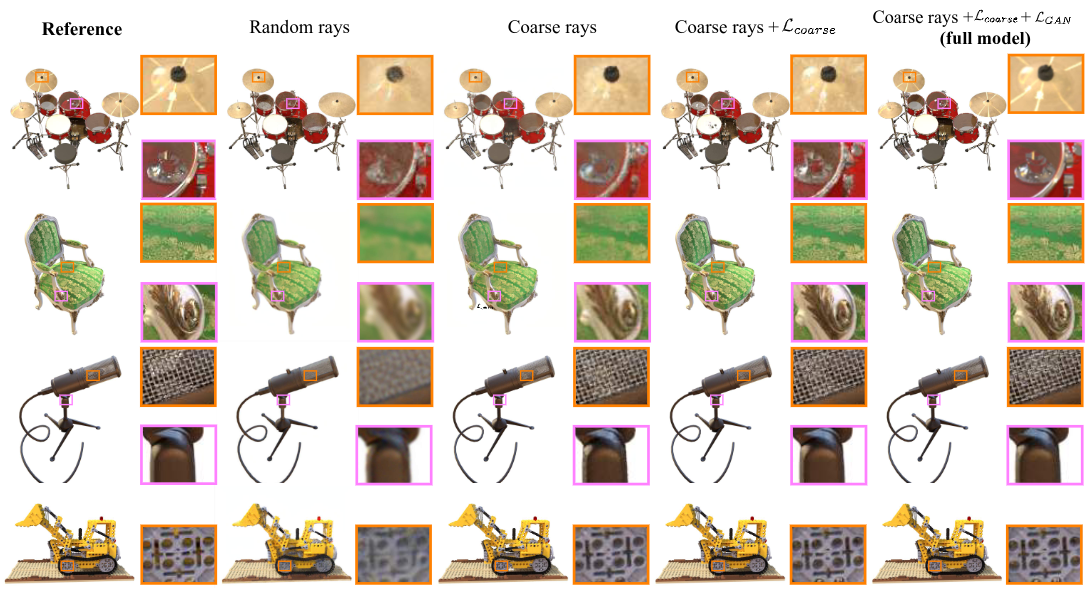}
  \caption{Qualitative ablation study. Comparison of ground-truth with predicted novel views by CG-NeRF with random and coarse rays training, with coarse reconstruction loss $\mathcal{L}_{coarse}$ and the full model with enabled adversarial loss $\mathcal{L}_{GAN}$. The full model not only predicts novel views more accurate than other baselines but also render them efficiently. Without using $\mathcal{L}_{GAN}$, we observe less realistic novel images compared to the ground-truths.}
\label{fig_abs}
\end{figure*}

\begin{table}[t]
\caption{CG-NeRF architecture ablation study. Reconstruction accuracy of view synthesis on the Forward-Facing scenes~\cite{DTU}.}
\centering
\resizebox{\columnwidth}{!}{%
\begin{tabular}{cccc|ccc}
\hline
Random & Coarse & $\mathcal{L}_{coarse}$ & $\mathcal{L}_{GAN}$ & PSNR$\uparrow$  & SSIM$\uparrow$ & LPIPS$\downarrow$ \\ \hline
\checkmark      &        &        &     & 19.15 & 0.82 & 0.34  \\
       & \checkmark       &        &     & 20.59 & 0.76 & 0.30  \\
       & \checkmark       & \checkmark       &     & 22.19 & 0.80 & 0.25  \\
       & \checkmark       & \checkmark       & \checkmark    & \textbf{23.93} & \textbf{0.82} & \textbf{0.21}  \\ \hline
\end{tabular}%
}
\label{tab_abs}
\end{table}

\section{Ablation Study}

\subsection{Architecture design}

Table~\ref{tab_abs} and Fig.~\ref{fig_abs} summarize the quantitative and qualitative results of CG-NeRF on different architectural choices using the test set of the Forward-Facing dataset \cite{TankAndTemples}. We first define a "Random rays" variant of CG-NeRF that estimates stochastic sampled pixels~\cite{wang2021ibrnet,chen2021mvsnerf} of the high-resolution novel views during training. Independently rendering each pixel leads to visible artifacts and blurriness of the predicted novel views. The rendering results are better if we estimate the entire the novel views using a convolution-based neural renderer. However, this model does not produce plausible target views as they still contain incorrect geometry and poorly rendered specular areas. By regularizing our model with a coarse reconstruction loss $\mathcal{L}_{coarse}$, we address the above issues and observe vastly improved novel views. Finally, we found that adding a hinge GAN loss $\mathcal{L}_{GAN}$ and the dual discriminator of~\cite{eff3DGAN} helps us to achieve state-of-the-art results as can be seen in the last column of Fig.~\ref{fig_abs}. We provide more comparison results in the supplementary videos.

\begin{table}[t]
\caption{The impact of the number of reference images, measured in terms of reconstruction accuracy on the Tank\&Temples~\cite{TankAndTemples}.
}
\begin{tabular}{@{}rccccccc@{}}
\toprule
      & \multicolumn{7}{c}{\# of reference images }                               \\ \cmidrule(l){2-8} 
      & 4     & 5     & 6     & 7              & 8     & 9     & 10     \\ \midrule
PSNR$\uparrow$ & 18.82 & 19.05 & 19.67 & \textbf{21.28} & 21.15 & 21.03 & 21.03 \\
LPIPS$\downarrow$ & 0.267 & 0.203 & 0.185 & \textbf{0.160} & 0.168 & 0.171 & 0.175 \\
SSIM$\uparrow$  & 0.796 & 0.825 & 0.871 & \textbf{0.892} & 0.890 & 0.890 & 0.887 \\ \bottomrule
\end{tabular}
\label{table_inputs}
\end{table}

\begin{figure*}[t]
\centering
  \includegraphics[width=0.99\linewidth]{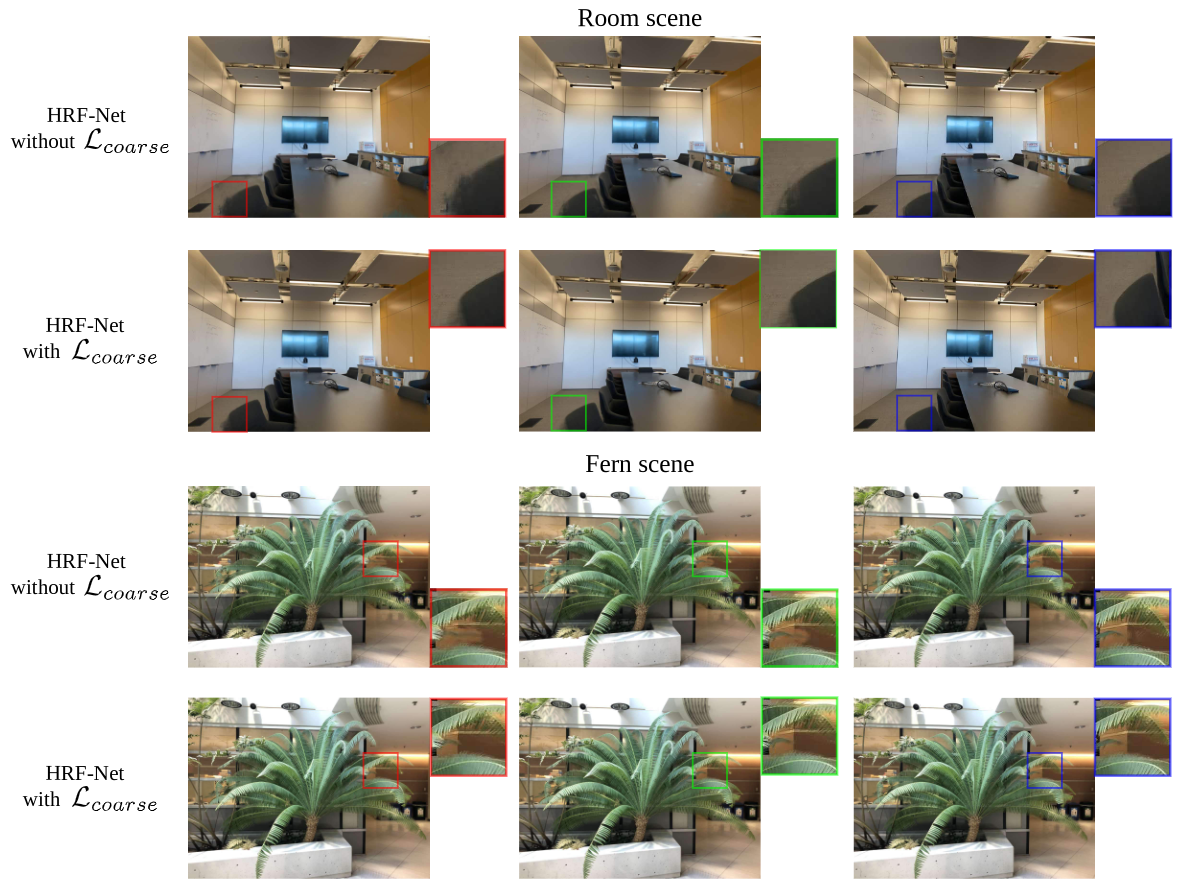}
  \caption{A generated sequence of consecutive novel views produced by CG-NeRF with and without the coarse reconstruction loss $L_{coarse}$ from the Room and Fern scenes of the LLFF~\cite{llff} dataset.}
\label{fig_seq}
\end{figure*}

\subsection{Rendering large 4K novel views}
Since we use a convolution-based neural renderer to obtain the high resolution images, CG-NeRF can accept very large 4K input images of the Forward-Facing dataset~\cite{llff} and generate new views.
In the Table~\ref{tab_improving}, we compare our approach with other variants of NeRF at both 800x800 and 4K resolution.

We first conduct an experiment to test whether if our neural renderer is able to improve existing view synthesis methods such as PointNeRF~\cite{pointnerf} and Instant-NGP~\cite{mueller2022instant} at the testing standard 800x800 resolution. Methods with the $^\ddagger$ symbol indicate that they were trained to produce a smaller novel views and then later up-sampled to the original size using our neural renderer. Given the same training time, both baseline methods produce almost similar novel views but the rendering time is significantly improved. This further highlights the usefulness of our novel neural renderer to the recently proposed methods that it can be easily plug into the existing systems.

We also found that directly optimizing both PointNeRF~\cite{pointnerf} and Instant-NGP~\cite{mueller2022instant} on 4K images requires more than 20GB of GPU memory and a longer training time to obtain good synthesis results. Therefore, we apply the same up-scaling strategy above to reduce the memory footprint. 
In contrast, our method only requires approximately 5GB of GPU memory to synthesis high quality 4K novel views. 
Instead of training $\text{CG-NeRF}^\ddagger$ model from scratch, we finetune our generalized CG-NeRF model and output 4K images. Experimental results in the Table~\ref{tab_improving} show that optimizing a convolution-based neural renderer improves the synthesis quality of the novel views at both testing standard 800x800 and very high 4K resolution.
Moreover, it takes 2.5 seconds to render a single 4K novel image but our method is still much faster compared to other baselines.

\subsection{Spatial-temporal consistency}

From the supplementary material, both of our generalized and finetuned CG-NeRF model renders photo-realistic high quality novel views with multi-view consistency using the learned encoder-decoder structure. 
Similar results have been observed in recent 3D generative NeRF-based methods~\cite{gu2022stylenerf,eff3DGAN} that we can produce high resolution 3D consistent novel views from 2D low resolution feature maps. 
In this work, we follow the design of the recently proposed EG3D~\cite{eff3DGAN} that uses a dual-discriminator to enforce consistent results between high and low resolution outputs. The up-sampling neural renderer of ~\cite{eff3DGAN} is similar to our proposed encoder-decoder network but we condition the synthesis process using a set of sparse input views.

As can be seen in the Fig.~\ref{fig_coarse_net}, we also use an aggregate warped features as input to the neural renderer. 
This warped features are consistent with the coarse radiance fields features since we leverage predicted coarse depth maps to perform feature fusion as described in the section~\ref{fine_net}. 
Therefore, we add a an regularization $L_{coarse}$ loss to train the coarse radiance fields predictor.
We found that using a simple and yet effective loss function
not only boosts the quality of the low and high resolution novel views but also improves the temporal consistency between consecutive novel views.
Without $L_{coarse}$, the predicted novel views include significant artifacts near the boundary and also not very temporally consistent due to independent renderings at each novel viewpoint using 2D Unet renderer (see Fig.~\ref{fig_seq}). 
By forcing the network to estimate accurate down-sampled novel views, our method can learn to produce consistent features in the higher resolution. We also try to the improved designs of~\cite{gu2022stylenerf} on our pipeline and observe similar results.

\subsection{Number of input views} 
In Table~\ref{table_inputs}, we evaluate the performance of our method with an increasing number of source images using the Tanks and Temples \cite{TankAndTemples} dataset. We report both SSIM and LPIPS metrics with the number of source images up to 10. We observe that CG-NeRF performs the best with 7 input views and then the results get worse. 
When reference and target poses are far from each other, inaccurate regressed depth maps will lead to less accurate novel views. Therefore, having views close to the target views and having less self-occlusion is essential to synthesize novel views. If it is hard to gather views around the target view, adding more views that have overlapping viewing frustums with the target view is also necessary.

\section{Conclusion}
We presented CG-NeRF, a new method to address the challenging problem of novel view synthesis from a sparse and unstructured set of input images.
Due to its coarse neural radiance field predictor and a convolution-based neural renderer, CG-NeRF can produce all pixels of the target view without relying any additional explicit data structure.  
Moreover, it enables highly efficient per-scene optimization that takes only 10-15 minutes, leading to rendering quality comparable to and even surpassing recent state-of-the-art methods which require several hours of training.

\bibliography{refs.bib}{}
\bibliographystyle{IEEEtran}

%
\begin{IEEEbiography}[{\includegraphics[width=1in,height=1.25in,clip,keepaspectratio]{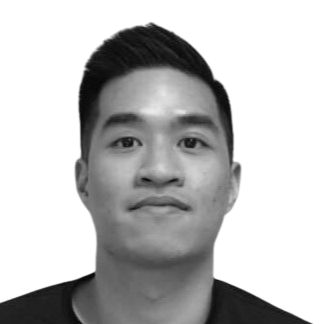}}]{Phong Nguyen-Ha} received the B.Ss. degree in mechanical engineering from the Ha Noi University of Science and Technology (HUST), Vietname and the M.Sc. degree in computer science engineering at Dongguk University, Seoul, South Korea. Since then, he is working as a doctoral candidate at CMVS, fully funded by the Vision-based 3D perception for mixed reality applications grant from the Infotech institute. His research interests include 3D computer vision, computer graphics and deep learning.

\end{IEEEbiography}

\begin{IEEEbiography}[{\includegraphics[width=1in,height=1.25in,clip,keepaspectratio]{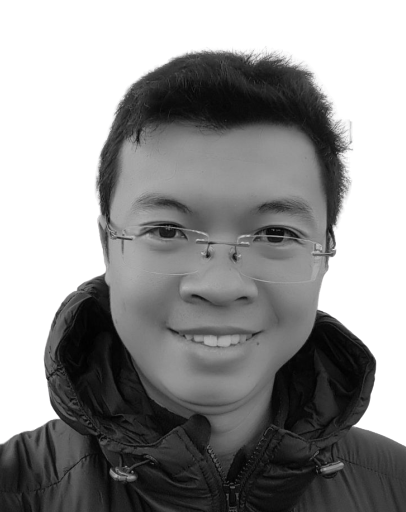}}]{Lam Huynh} received the B.Ss. degree in computer science from the University of Information Technology, Ho Chi Minh city, Vietnam and the M.Sc. degree in CSE at the Center for Machine Vision and Signal Analysis University of Oulu. Since then, he is working as a doctoral candidate at CMVS, fully funded by the Vision-based 3D perception for mixed reality applications grant from the Infotech institute. His research interests include computer vision, deep learning, 3D sensing and 3D scene understanding.

\end{IEEEbiography}

\begin{IEEEbiography}[{\includegraphics[width=1in,height=1.25in,clip,keepaspectratio]{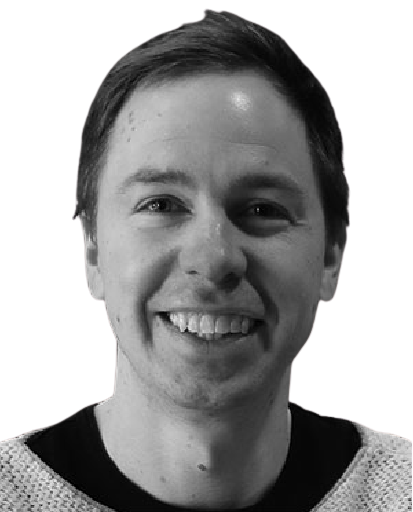}}]{Esa Rahtu} received his PhD degree from the University of Oulu in 2007. Currently he is an Assistant Professor at Tampere University of Technology (TUT) in Finland. Prior to joining TUT, Rahtu was a senior researcher at the Center of Machine Vision research at the University of Oulu in Finland. In 2008, he was awarded a post-doctoral research fellow funding by the Academy of Finland. His main research interests are in computer vision and deep learning.

\end{IEEEbiography}

\begin{IEEEbiography}[{\includegraphics[width=1in,height=1.25in,clip,keepaspectratio]{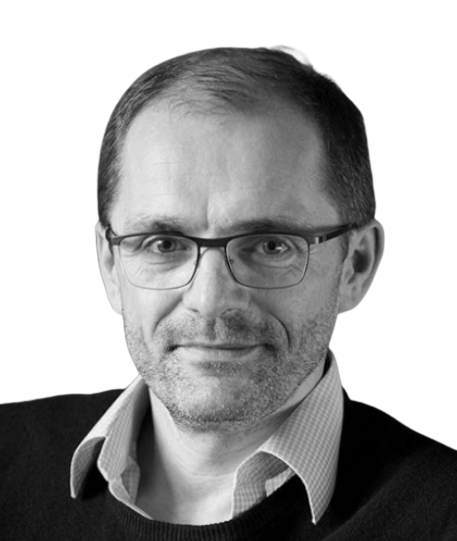}}]{Jiri Matas} received the Ph.D. degree from the University of Surrey, Guildford, U.K., in 1995. He is a Professor with Center for Machine Perception, Czech Technical University, Prague, Czech Republic. He has published more than 200 articles in refereed journals and conferences. His publications have approximately 34,000 citations in Google Scholar and 13,000 in the Web of Science. His H-index is 65 (GS) and 43 (WoS), respectively. His research interests include visual tracking, object recognition, image matching and retrieval, sequential pattern recognition, and RANSAC-type optimization methods. He is on the editorial board of the IJCV and was an Associate Editor-in-Chief of the IEEE TRANSACTIONS ON PATTERN ANALYSIS AND MACHINE INTELLIGENCE (TPAMI).

\end{IEEEbiography}

\begin{IEEEbiography}[{\includegraphics[width=1in,height=1.25in,clip,keepaspectratio]{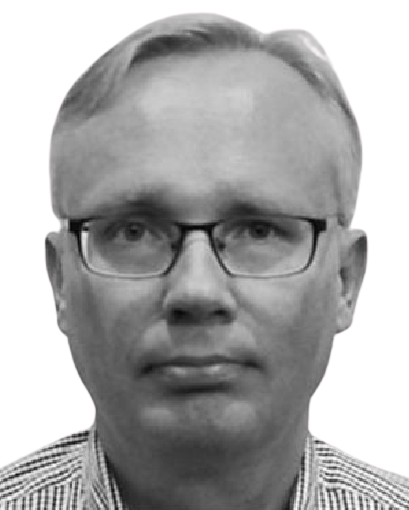}}]{Janne Heikkil\"a} received the Doctor of Science in Technology degree in information engineering from the University of Oulu, Oulu, Finland, in 1998. He is currently a Professor of computer vision and digital video processing with the Faculty of Information Technology and Electrical Engineering, University of Oulu, and the Head of the Degree Program in computer science and engineering. He has supervised nine completed doctoral dissertations and authored/coauthored more than 160 peer-reviewed scientific articles in international journals and conferences. His research interests include computer vision, machine learning, digital image and video processing, and biomedical image analysis.Prof. Heikkilä has served as an Area Chair and a member of program and organizing committees of several international conferences. He is a Senior Editor for the Journal of Electronic Imaging, an Associate Editor for the IET Computer Vision and Electronic Letters on Computer Vision and Image Processing, a Guest Editor for a special issue in Multimedia Tools and Applications, and a member of the Governing Board of the International Association for Pattern Recognition. During 2006–2009, he was the President of the Pattern Recognition Society of Finland. He has been the Principal Investigator in numerous research projects funded by the Academy of Finland and the National Agency for Technology and Innovation.

\end{IEEEbiography}




\end{document}